\documentclass{article}

\usepackage{xcolor}
\usepackage[preprint]{corl_2026} 

\usepackage{amsmath}
\usepackage{booktabs}
\usepackage{graphicx}
\usepackage{capt-of}
\usepackage{flafter} 
\usepackage{tikz}
\usetikzlibrary{positioning,fit,backgrounds,arrows.meta}

\newcommand{\wm}{Cosmos-Predict1-7B Video2World}

\title{Dreams Are Cheaper Than Flights: Curated World-Model Data Improves Aerial Detection}

\author{
  Rugved Katole \quad Vedant Patil \quad Jenna Kline \quad Christopher Stewart\\
  Department of Computer Science and Engineering\\
  The Ohio State University, Columbus, OH, USA\\
  \texttt{katole.2@osu.edu, cstewart@cse.ohio-state.edu}
}

\begin{document}
\maketitle
\suppressfloats[t]


\begin{abstract}
Robot perception is increasingly data driven, yet field data is slow, costly, and sometimes disruptive to collect.
Video world models offer a cheaper source: seeded with real frames, they extend a recorded scene into new footage with the deployment's sensor, altitude, and environment.
We ask whether such synthetic data matches the utility of real field data and whether detectors trained with it perform in the field.
Seeding a 7B-parameter world model with wildlife drone video, we more than double the training set with curated frames.
On held-out videos from the same drones, mAP@50--95 rises 4.3\% when these frames are added to real data and 10.3\% when they replace it, and up to 33.5\% on an urban benchmark.
Although an expert rejects an estimated 87.5\% of generated clips, the data costs about \$1,100 of GPU time, against more than \$3,200 of field labor alone for a less useful real training set.
Synthetic data thus shifts data costs from fieldwork to compute.
\end{abstract}



\section{Introduction}
\label{sec:intro}

Modern robot perception is trained rather than engineered, and its reliability tracks the quantity and diversity of the data behind it.
For field robots this dependence is expensive.
Autonomous drones that detect and follow animal herds for ecological monitoring~\cite{kline2026wildwing,kline2025edge,kline2025collective} need low-altitude video of the species, sites, and behaviors they will encounter, collected within permits, weather windows, and flight-safety rules~\cite{maalouf2026sora} and without disturbing the animals under study~\cite{afridi2025disturbance}.
The MMLA dataset required more than 150 person-hours of fieldwork on two continents to record 37 videos~\cite{kline_mmla_2025}, and comparable campaigns underlie other wildlife drone datasets~\cite{kholiavchenko2024kabr,kline2025smartwilds}.
Even so, such datasets cover few herds, days, and poses, and models trained in such small-data regimes generalize poorly~\cite{stevens2025datagap}.

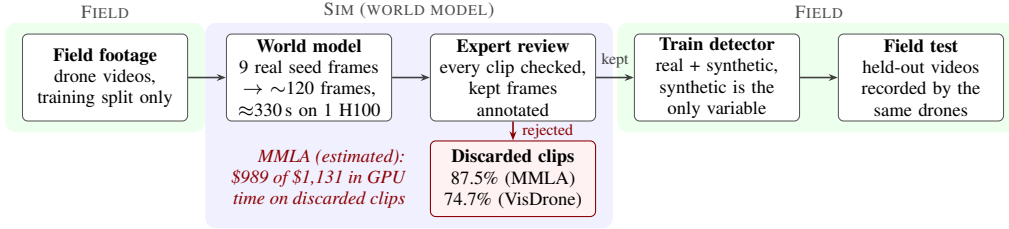
\begin{figure}[t]
\centering
\begin{tikzpicture}[
  font=\scriptsize,
  node distance=2.5mm and 5mm,
  box/.style={draw=black!65, rounded corners=2pt, align=center, text width=21mm,
              minimum height=11.5mm, inner sep=1.2pt, fill=white},
  bad/.style={box, draw=red!55!black, fill=red!5, minimum height=10mm},
  arr/.style={-{Stealth[length=3.5pt]}, semithick, black!75},
  band/.style={rounded corners=4pt, inner xsep=2.2mm, inner ysep=1.4mm},
  bandlabel/.style={font=\scriptsize\scshape, text=black!70, anchor=south, inner sep=1.5pt}
]
  \node[box] (field) {\textbf{Field footage}\\drone videos,\\training split only};
  \node[box, right=of field] (wm) {\textbf{World model}\\9 real seed frames\\$\rightarrow$ ${\sim}$120 frames,\\$\approx$330\,s on 1 H100};
  \node[box, right=of wm] (rev) {\textbf{Expert review}\\every clip checked,\\kept frames\\annotated};
  \node[box, right=of rev] (train) {\textbf{Train detector}\\real + synthetic,\\synthetic is the\\only variable};
  \node[box, right=of train] (test) {\textbf{Field test}\\held-out videos\\recorded by the\\same drones};
  \node[bad, below=of rev] (drop) {\textbf{Discarded clips}\\87.5\% (MMLA)\\74.7\% (VisDrone)};
  \node[left=2mm of drop, align=right, text=red!55!black, font=\scriptsize\itshape] (waste)
       {MMLA (estimated):\\\$989 of \$1,131 in GPU\\time on discarded clips};

  \draw[arr] (field) -- (wm);
  \draw[arr] (wm) -- (rev);
  \draw[arr] (rev) -- node[above, font=\tiny] {kept} (train);
  \draw[arr] (train) -- (test);
  \draw[arr, red!55!black] (rev) -- node[right, font=\tiny] {rejected} (drop);

  \begin{scope}[on background layer]
    \node[band, fill=green!7, fit=(field)] (b1) {};
    \node[band, fill=blue!6, fit=(wm)(rev)(drop)(waste)] (b2) {};
    \node[band, fill=green!7, fit=(train)(test)] (b3) {};
  \end{scope}
  \node[bandlabel] at (b1.north) {Field};
  \node[bandlabel] at (b2.north) {Sim (world model)};
  \node[bandlabel] at (b3.north) {Field};
\end{tikzpicture}
\caption{Field-to-simulation-to-field protocol: training footage seeds a world model, an expert discards deformed clips and annotates the rest, and detectors are tested on held-out field videos.}
\label{fig:pipeline}
\end{figure}

Simulation is robotics' long-standing answer to scarce data.
Rendered simulators make data cheap but must cross a reality gap, typically by randomizing appearance until the real world looks like one more variation~\cite{tobin2017domain}.
This presumes a simulator that can render the target, which is hard for animals in unstructured terrain.
Generative models offer a different trade.
Diffusion-generated images improve recognition with scarce real data~\cite{he2023synthetic,trabucco2023effective} and augment robot learning data~\cite{chen2023genaug,yu2023rosie}, flow-based interpolation densifies sparse aerial imagery~\cite{katole2025orthofuse}, and video world models roll a scene forward from a few real frames~\cite{ha2018world,yang2024unisim,bruce_genie_2024,nvidia_cosmos_2025}.
Seeded with field footage, such a model yields clips that inherit the sensor, altitude, and environment of the source, which we call \emph{field-seeded} data.

Generated data is not free of risk: world models still produce visible deformities, such as duplicated heads (Fig.~\ref{fig:failures}), and models trained on generated data can lose the tails of the real distribution~\cite{shumailov2024collapse}.
Synthetic clips must therefore be curated, and because curation follows generation, every rejected clip has already consumed its full compute budget.
This raises a practical question rarely answered with numbers: despite high rejection rates, is synthetic data still cheaper than returning to the field?

This paper asks whether synthetic data generated from field seeds can provide the fidelity and accuracy of real field data, and whether detectors trained with it perform well on real field footage.
We answer both with a field-to-simulation-to-field protocol in which a world model is seeded only from training videos, an expert curates its output, and detectors are tested on held-out videos from the same drones, so that the training data is the only variable.
We also price it against fieldwork.


\section{Methodology}
\label{sec:method}

Our protocol begins and ends in the field (Fig.~\ref{fig:pipeline}).
We seed a world model with training footage, curate its output, and test detectors trained with and without it on held-out video from the same drones.
We generate with \wm{}~\cite{nvidia_cosmos_2025}, an open world foundation model that continues a video from conditioning frames.
Each generation is seeded with nine consecutive training frames and a short text prompt describing the scene, and produces a clip of about 120 frames under the model's default sampler settings.
Because the seed fixes the camera, altitude, terrain, and illumination, the model invents how the scene evolves, not what it looks like, which avoids most of the appearance gap that rendered simulators close by randomization~\cite{tobin2017domain}.

Generated clips fail in ways that a domain expert recognizes at once but a detector would absorb as features: distorted or duplicated anatomy, bodies or vehicles that merge or vanish, and motion that violates physics (Fig.~\ref{fig:failures}).
A human expert therefore reviews every clip, rejects any clip with a deformity, and annotates the frames of accepted clips with bounding boxes.
Only accepted frames enter training.
Curation turns generation into rejection sampling.
If a clip costs $T$ GPU-seconds to generate and is accepted with probability $p$, the expected compute per accepted clip and the fraction of compute spent on rejected clips are
\begin{equation}
  T_{\text{usable}} = T/p, \qquad f_{\text{waste}} = 1 - p .
  \label{eq:cost}
\end{equation}
The first quantity scales with $T$, which diffusion acceleration reduces.
The second is set by the world model's failure rate in the target domain, which acceleration leaves untouched.

We apply this protocol to two drone datasets.
MMLA~\cite{kline_mmla_2025} contains 155,074 annotated frames from 37 low-altitude videos recorded at Ol Pejeta and Mpala in Kenya and The Wilds in Ohio with DJI Air~2S, DJI Mini, and Parrot Anafi drones.
Following the benchmark protocol, we split it into training and test partitions and subsample the training partition to 20,618 frames with 85,967 boxes of zebra, giraffe, and onager.
The test partition is never used for seeding, prompting, or curation.
To test generality beyond wildlife, we repeat the protocol on VisDrone~\cite{zhu2021visdrone}: 24,201 training frames of urban drone footage, tested on its official validation split (2,846 frames, 114,132 boxes, ten classes).
On MMLA, the expert reviewed every generated clip, and accepted clips yielded 26,702 synthetic frames with 93,004 boxes.
The combined training set thus has 2.3$\times$ the frames and 2.1$\times$ the boxes of the real one.
In an expert-labeled 40-clip pilot, five clips passed ($p = 0.125$).
On VisDrone, the expert reviewed 300 clips in 45 minutes and accepted 76 ($p = 0.253$, 9,204 frames).
The 224 rejected clips (27,896 frames) were annotated the same way and serve as an uncurated control.

On MMLA we fine-tune YOLO11m~\cite{jocher2024yolo11} for 50 epochs at batch size 96 with the benchmark's hyperparameters.
On VisDrone we train YOLOv8m~\cite{jocher2023yolov8} for 80 epochs at batch size 64.
Within each dataset the detector, hyperparameters, and test split are fixed, so the training data is the only variable.
We report mAP@50--95 on the field test split.
Generation ran on one NVIDIA H100 (80\,GB HBM3, 116\,GB system memory) at about 330\,s per clip, priced at \$6.93 per GPU-hour, the mean October 2026 on-demand rate of H100 instances on AWS (p5.4xlarge) and Azure (NC40ads H100 v5).


\section{Results}
\label{sec:results}

\begin{figure}[htb]
\begin{minipage}[b]{0.49\linewidth}
  \centering
  \captionof{table}{Generation cost at \$6.93/H100-hour, excluding annotation and training. $^\dagger$Estimated from the pilot.}
  \label{tab:cost}
  \vspace{4pt}
  \scriptsize
  \setlength{\tabcolsep}{3pt}
  \begin{tabular}{lrr}
    \toprule
                          & MMLA                                & VisDrone \\
    \midrule
    Pass rate $p$         & 12.5\%                              & 25.3\% \\
    Frames gen.\ / kept   & 213.6K$^\dagger$ / 26.7K            & 37.1K / 9.2K \\
    GPU, total            & 163.2\,h / \$1,131$^\dagger$        & 27.5\,h / \$191 \\
    GPU, discarded        & 142.8\,h / \$989$^\dagger$          & 20.5\,h / \$142 \\
    \$ per 1k kept frames & \$42.4$^\dagger$                   & \$20.7 \\
    \bottomrule
  \end{tabular}
\end{minipage}\hfill
\begin{minipage}[b]{0.48\linewidth}
  \centering
  \includegraphics[width=\linewidth]{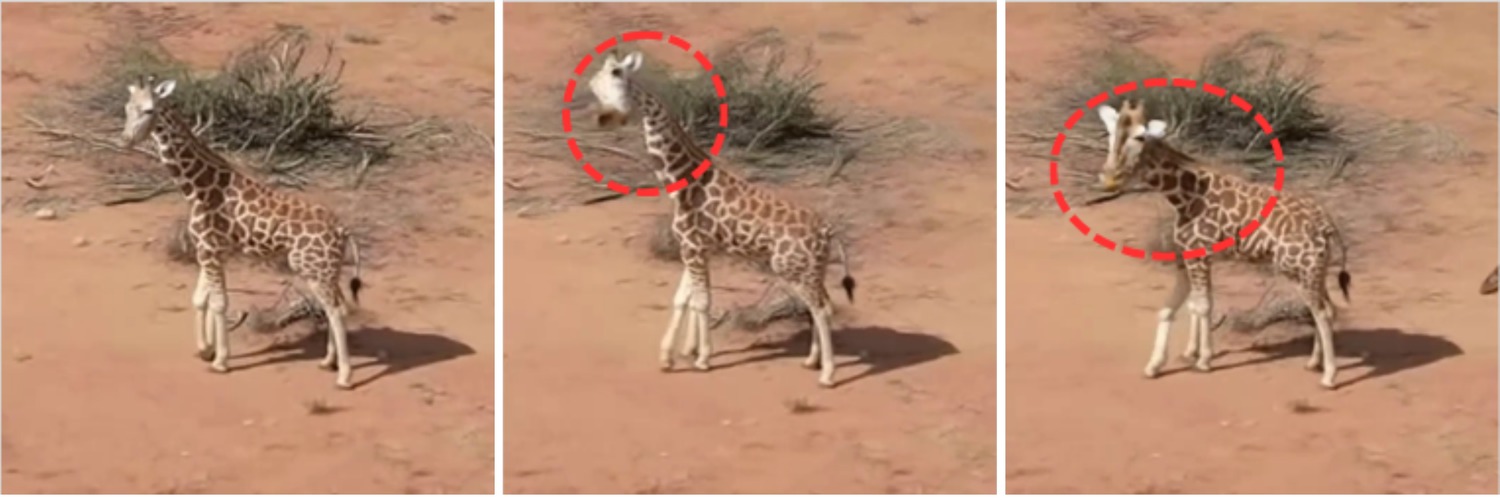}
  \captionof{figure}{A rejected generation: seed frame (left) and generated frames with a distorted head and neck (circled).}
  \label{fig:failures}
\end{minipage}
\end{figure}

We evaluate both datasets with the same training study.
A detector trained on real field frames alone sets the baseline.
We then add increasing fractions of the curated pool to the full real training set and track field accuracy and localization.
On MMLA we also train on synthetic frames alone, and on VisDrone on real plus rejected frames.
The synthetic data has a measurable price (Table~\ref{tab:cost}).
On MMLA, the 26,702 retained frames more than double the training set.
At the pilot rejection rate of 87.5\%, obtaining them required about 213.6K generated frames and 163 H100-hours (\$1,131).
Of this, 143 hours (\$989) produced discarded clips, so an accepted clip costs \$5.08 rather than \$0.64.
VisDrone's acceptance rate is twice as high, yet three quarters of its \$191 still pays for rejected clips.

\subsection{Detection on Held-Out Field Video}

Table~\ref{tab:mmla} reports the MMLA training study.
mAP@50--95 rises with every increase in synthetic volume, from 0.845 with real data alone to 0.881 (+4.3\%).
That is about $+0.0013$ per 1,000 synthetic frames, or per \$42 of GPU time.
Meanwhile mAP@50 stays near 0.98, so synthetic data tightens localization rather than finding more animals.
Synthetic frames alone score highest (0.932, +10.3\%), beating the real-only detector on every class (zebra 0.787 to 0.939, onager 0.854 to 0.950, giraffe 0.894 to 0.907).
The test videos were never used as seeds, so this is not leakage.
Instead, it is consistent with three properties of the curated pool, which single runs cannot separate.
First, the pool is larger (29.5\% more frames, 8.2\% more boxes) and extends each seed into new poses.
Second, generated frames tend to carry less motion blur and sensor noise than drone footage, which may sharpen object boundaries.
Third, one expert annotated the whole pool in a single pass, whereas the real labels come from a large multi-annotator effort~\cite{kline_mmla_2025}.
All three favor strict-IoU localization, which is exactly where the synthetic-only detector gains.
Its mAP@50, precision, and recall stay at or below the real-only values, and mixing both sources (0.881) lands in between.

\begin{table}[htbh]
\centering
\caption{Field results on MMLA (YOLO11m, held-out field test). ``Real + $x$\%'' adds $x$\% of the curated synthetic pool to the 20,618 real frames. $\Delta$ is relative to real only.}
\label{tab:mmla}
\footnotesize
\begin{tabular}{lrrcccl}
\toprule
Training data      & Synthetic & Total  & P              & R              & mAP@50         & mAP@50--95 ($\Delta$) \\
\midrule
Real only          & 0         & 20,618 & \textbf{0.983} & \textbf{0.945} & \textbf{0.984} & 0.845 \\
Real + 30\% syn.   & 8,010     & 28,628 & 0.970          & \textbf{0.945} & 0.981          & 0.852 (+0.8\%) \\
Real + 40\% syn.   & 10,680    & 31,298 & 0.970          & 0.937          & 0.978          & 0.856 (+1.3\%) \\
Real + 50\% syn.   & 13,351    & 33,969 & 0.970          & 0.936          & 0.978          & 0.859 (+1.7\%) \\
Real + 60\% syn.   & 16,021    & 36,639 & 0.970          & 0.935          & 0.976          & 0.862 (+2.0\%) \\
Real + 100\% syn.  & 26,702    & 47,320 & 0.970          & 0.930          & 0.979          & 0.881 (+4.3\%) \\
\midrule
Synthetic only     & 26,702    & 26,702 & 0.965          & 0.939          & 0.981          & \textbf{0.932} (+10.3\%) \\
\bottomrule
\end{tabular}
\end{table}

Table~\ref{tab:visdrone} reports the VisDrone study.
Curated frames raise mAP@50--95 from 0.138 to 0.184 (+33.5\%) at 75\% of the pool, with the largest gains on rarer vehicles (truck 0.050 to 0.277, van 0.086 to 0.195, car 0.374 to 0.387).
Rejected frames help almost as much and give the best van and truck AP.
Rejection is conservative: one deformed object discards a whole clip.
VisDrone scenes hold about 40 boxes per frame, and a deformity typically affects one or two vehicles, so most boxes in rejected clips stay valid.
The rejected pool is also three times larger.
Yet curated data reaches the same aggregate with a quarter of the synthetic frames (6,905 versus 27,896).
The damage also concentrates where valid examples are scarce: bus, the rarest class, falls from 0.211 to 0.010.
Curation thus buys efficiency and limits rare-class damage that aggregate accuracy hides.

\begin{table}[htb]
\centering
\caption{Field results on VisDrone (YOLOv8m, validation split). ``+ $x$\% curated'' adds a nominal $x$\% of the 9,204 kept frames, and ``+ rejected'' adds the 27,896 rejected ones. $\Delta$ is relative to real only and computed from unrounded scores.}
\label{tab:visdrone}
\footnotesize
\setlength{\tabcolsep}{4pt}
\begin{tabular}{lrlcccc}
\toprule
                  &           &                          & \multicolumn{4}{c}{Per-class AP@50--95 (test instances)} \\
\cmidrule(lr){4-7}
Training data     & Synthetic & mAP@50--95 ($\Delta$)    & Car (31,821)   & Van (6,842)    & Truck (1,359)  & Bus (264) \\
\midrule
Real only         & 0         & 0.138                    & 0.374          & 0.086          & 0.050          & \textbf{0.211} \\
+ 25\% curated    & 2,307     & 0.167 (+20.6\%)          & \textbf{0.388} & 0.135          & 0.184          & 0.160 \\
+ 50\% curated    & 4,606     & 0.175 (+26.9\%)          & 0.368          & 0.157          & 0.209          & 0.200 \\
+ 75\% curated    & 6,905     & \textbf{0.184} (+33.5\%) & 0.387          & 0.195          & 0.277          & 0.130 \\
+ 100\% curated   & 9,204     & 0.182 (+31.5\%)          & 0.382          & 0.174          & 0.284          & 0.147 \\
\midrule
+ rejected        & 27,896    & 0.183 (+32.5\%)          & 0.375          & \textbf{0.196} & \textbf{0.330} & 0.010 \\
\bottomrule
\end{tabular}
\end{table}

\subsection{Synthetic Generation Versus Field Collection}

Despite the waste, synthetic data is cheaper than the fieldwork it replaces.
The MMLA campaign took more than 150 person-hours on two continents~\cite{kline_mmla_2025}, at least \$5,500 at the 2025 median wage of U.S. wildlife biologists (\$36.91/h~\cite{bls2025wildlife}), before airfare, lodging, permits, flight approvals~\cite{maalouf2026sora}, and equipment.
Our real training set comes from the training partition, 58\% of the data~\cite{kline_mmla_2025}, so its share of field labor is about 87 person-hours (\$3,210).
The synthetic pool, which alone trained a more accurate field detector, cost \$1,131 of GPU time plus about \$164 of expert review (about 1,780 clips at 9\,s each, the same rate as on VisDrone).
We exclude annotation, which both routes need.
Even pricing fieldwork at labor alone, synthetic data is 2.5 times cheaper.
It would be about 20 times cheaper without failed generations, since retained clips used only \$141 of GPU time.


\section{Conclusion}
\label{sec:conclusion}

Field-seeded world models can stand in for part of the field.
Seeded with drone footage, a world model produced frames that trained a better MMLA field detector than real frames and improved field accuracy on both datasets when added to real data.
Even with an estimated 87.5\% of clips discarded, this data cost about \$1,100 of GPU time.
That is less than the field labor behind a less useful real training set.
Two conditions qualify the result.
Synthetic output must be curated, and its transfer verified per class, since a poorly rendered class can vanish unnoticed in aggregate accuracy.


\section{Future Work}
\label{sec:future}

The dominant cost of field-seeded data is waste.
An estimated 87.5\% of MMLA generation compute (\$989 of \$1,131) produced clips that no detector saw.
Faster solvers~\cite{song_denoising_2021,lu_dpm-solver++_2023}, distillation~\cite{salimans_progressive_2022}, and within-network early exits~\cite{moon_simple_2024,tang_adadiff_2024} shrink $T$ in Eq.~\eqref{eq:cost} for every generation alike but leave $1-p$ unchanged.
Recovering this compute calls for quality-aware generation that stops failing generations early, with automated checks to relieve the expert.
Building such a system is our ongoing work.


\clearpage
\acknowledgments{Omitted for anonymous review.}

\bibliography{references}

\end{document}